# Data-Based MHE for Agile Quadrotor Flight

Wonoo Choo and Erkan Kayacan

*Abstract*— This paper develops a data-based moving horizon estimation (MHE) method for agile quadrotors. Accurate state estimation of the system is paramount for precise trajectory control for agile quadrotors; however, the high level of aerodynamic forces experienced by the quadrotors during high-speed flights make this task extremely challenging. These complex turbulent effects are difficult to model and the unmodelled dynamics introduce inaccuracies in the state estimation. In this work, we propose a method to model these aerodynamic effects using Gaussian Processes which we integrate into the MHE to achieve efficient and accurate state estimation with minimal computational burden. Through extensive simulation and experimental studies, this method has demonstrated significant improvement in state estimation performance displaying superior robustness to poor state measurements.

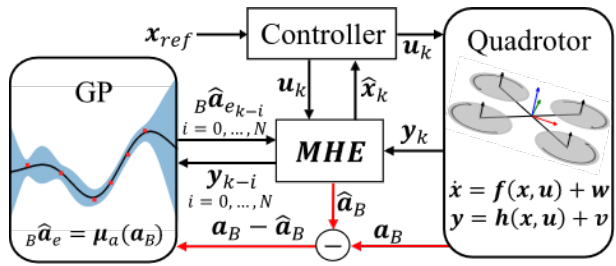

Fig. 1: Architecture of the GP based MHE. The offline computations are denoted in red and online computations are denoted in black.

## I. INTRODUCTION

The agility of the quadrotors makes them an ideal platform for many tasks and applications that require high speeds and maneuverability, such as explorations, inspections, light shows, drone filming, or search and rescue missions [1–5]. The ability to execute high-speed agile flights make quadrotors so versatile and desirable for many applications; however, acquiring the full state information of the system is a challenging task which is further encumbered by their restricted payload capacity [1, 6] Quadrotors are generally equipped with a GPS and an IMU. This combination of sensors measures the position, angular velocity and the linear acceleration of the system requiring the angular position and the linear velocity to be estimated. Moreover, due to the limited capability of the sensors, the measured states are also required to be estimated to handle the measurement noise.

The difficulties of designing accurate state estimator lies in modeling the dynamics of the system. The aerodynamic effects that the quadrotors experience during agile flights are challenging to model as they are a combination of propeller lift, interaction between the propellers, and drag from rotors and the fuselage [7, 8]. Furthermore, the model accuracy and complexity that can be utilized are constrained by the limited computing capabilities of the onboard executing platform. Using a kinematic model can be an advantageous option over a dynamic model due to the high-levels of unmodelled aerodynamic effects. However, low quality sensors can degrade the performance of kinematic model based estimator [9]. Dynamic model based estimator with an accurate model can be more robust to the poor state measurements and estimate unknown system parameters online, such as the payload mass, which can improve the state estimation and trajectory tracking performance [10–12].

For nonlinear systems, the most commonly used state estimation method is Extended Kalman Filter (EKF) [1, 10]. However, the highly nonlinear behaviors of agile quadrotors can lead to poor performance as EKF linearizes the nonlinear model around the current state estimates using Taylor expansions [3, 13]. Authors of [14] utilizes Convolutional Neural Network (CNN) to learn the IMU kinematic properties and the dynamics of the quadrotor to be integrated with the EKF to improve the state estimation. Though, this method improves the performance of the EKF, it requires large dataset of ground truth measurements to train the CNN which can be difficult to obtain [14].

Moving Horizon Estimation (MHE) is an alternative option to the EKF that is able to handle nonlinear models and system inequalities without linearization, which distinguishes it from other state estimation methods [11, 15]. MHE is an optimization-based method that utilizes the system model and the past measurements to estimate the system states. It is robust to poor initial guesses and guarantees local stability in contrast to EKF which cannot guarantee any general convergence and may fail with poor initialization [10, 15].

MHE and Model Predictive Control (MPC) are often referred to as the dual optimization problem of each other as they share the same optimal control structure [3, 16]. Authors of [7] incorporated Gaussian Processes (GP) to a simple nominal model of the system to improve the closed loop tracking control of the quadrotor by enhancing the dynamic model of the system. The GP models were trained to learn the unmodelled aerodynamic effects of the system that are challenging to model. As performance of the MHE is also heavily dependent on the accuracy of the system model, we propose extending this approach to state estimation methods.

In this paper, we develop an MHE pipeline augmented with GP regression models that has learned the residual

Wonoo Choo and Erkan Kayacan are with the School of Aerospace and Mechanical Engineering, The University of Oklahoma, Norman, OK 73019, USA. Email: {w.choo, erkan}@ou.edu

dynamics using the onboard sensor measurements. We extend the data-driven approach for MPC presented in [7] to MHE by combining a simple dynamic model with GPs to improve the state estimation performance of agile quadrotors. This method significantly reduces the learning problem by only considering the unmodelled dynamics of the system and the data collection process is substantially simplified by removing the requirement of ground truth odometry. Moreover, the additional computation from the GPs are minimized by only requiring single point predictions from the current state measurements at each time step. We conduct comparative simulation and experimental studies of the developed method in this paper to MHEs with a kinematic model and a dynamic model with and without an unknown system parameter at varying noise levels on the measurements. At higher noise levels, the GP augmented MHE displayed improved robustness to poor state measurements over other MHEs with the added capability of estimating varying payload mass.

The paper is organized as follows: The dynamic model of the quadrotor is presented in Section II. The formulation of the MHEs investigated in this study are described in Section III. The traditional formulation of GP and the developed methodology for data collection and model learning is presented in Section IV. The simulation studies are presented in Section V and the experimental studies are presented in Section VI. Then finally, a brief conclusion is drawn in Section VII.

## II. PRELIMINARIES

### A. Notation

In this paper, we denote the scalars with lowercase, vectors with bold lowercase, and matrices with bold uppercase. The Euclidean vector norm is denoted as $\|\cdot\|$. The World and Body frame axis are shown in Fig. 2. The Body frame is at the center of mass of the quadrotor and the rotors are assumed to be on the $xy$-plane of the Body frame. A vector $\boldsymbol{v}$ pointing from $\boldsymbol{p}_1$ to $\boldsymbol{p}_2$ in the World frame is denoted as $_W\boldsymbol{v}_{12}$. If $\boldsymbol{p}_1$ is the origin of the frame it is described in, then the pre-subscript is omitted. The orientation of the quadrotor is represented using a unit quaternion, $\boldsymbol{q}_{WB} = (q_w, q_x, q_y, q_z)$ where $\|\boldsymbol{q}_{WB}\| = 1$. The quaternion frame rotation is given by the quaternion-vector product, $\odot$, such that $\boldsymbol{q} \odot \boldsymbol{v} = \boldsymbol{q}\boldsymbol{v}\bar{\boldsymbol{q}}$, where $\bar{\boldsymbol{q}}$ is the conjugate of $\boldsymbol{q}$.

### B. Quadrotor Dynamics

The quadrotor is modeled as a 6 degree-of-freedom rigid body with mass $m$. The quadrotor's states position, orientation, linear velocity and angular velocity is denoted as $\boldsymbol{x} = [\boldsymbol{p}_{WB}, \boldsymbol{q}_{WB}, \boldsymbol{v}_{WB}, \boldsymbol{\omega}_B]^\mathsf{T}$. The control input is given by the collective thrust $\mathrm{f}_{thrust}$ and the linear acceleration in body frame, $\boldsymbol{a}_B$ is given by (1).

$$\boldsymbol{a}_B = \begin{bmatrix} 0 \\ 0 \\ \mathrm{f}_{thrust}/m \end{bmatrix} \quad (1)$$

Here the angular velocity dynamics were ignored as the individual thrusts of the rotors were unknown. The dynamics of the angular velocity was set as 0 to be formulated as a measurement state in the MHE. The nonlinear state-space model of the system is 13-dimensional, and its dynamics is given by (2) where $\boldsymbol{g}_W$ denotes the Earth's gravity.

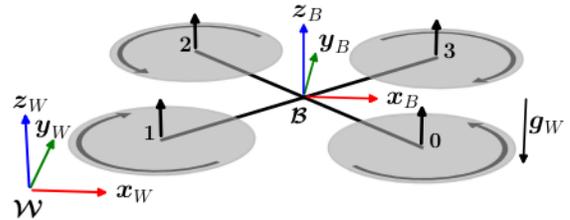

Fig. 2: Visualization of the quadrotor model and its reference frames.

$$\dot{\boldsymbol{x}} = \boldsymbol{f}_{dyn}(\boldsymbol{x}, \boldsymbol{u}) = \begin{bmatrix} \boldsymbol{v}_{WB} \\ \boldsymbol{q}_{WB} \cdot \begin{bmatrix} 0 \\ \boldsymbol{\omega}_B/2 \end{bmatrix} \\ \boldsymbol{q}_{WB} \odot \boldsymbol{a}_B + \boldsymbol{g}_W \\ 0 \end{bmatrix} \quad (2)$$

## III. MHE FORMULATION

MHE is an optimization-based state estimation method that utilizes the system model and the past measurements [13]. Theoretically, the MHE solves for an infinite horizon optimization problem [10, 11]. However, as this is computationally intractable in real-time, MHE reconciles only a finite number of recent measurements in the estimation horizon of length $N$ and the measurements collected before the estimation horizon are summarized by the arrival cost $\bar{\boldsymbol{x}}_{k-N}$ [10, 13, 15]. The MHE problem is formulated as follows:

$$\min_{\boldsymbol{x}, \boldsymbol{w}} \sum_{i=k-N}^{k} \|\boldsymbol{y}_i - \hat{\boldsymbol{y}}_i\|^2_{\boldsymbol{R}_k^{-1}} + \|\boldsymbol{w}_i\|^2_{\boldsymbol{Q}_k^{-1}} \quad (3a)$$
$$+ \|\boldsymbol{x}_{k-N} - \bar{\boldsymbol{x}}_{k-N}\|^2_{\boldsymbol{Q}_{k-N}^{-1}}$$

subject to $\quad \boldsymbol{x}_{k+1} = \boldsymbol{f}_{RK4}(\boldsymbol{x}_k, \boldsymbol{u}_k) + \boldsymbol{w}_k \quad (3b)$
$$\boldsymbol{y}_k = \boldsymbol{h}(\boldsymbol{x}_k, \boldsymbol{u}_k) + \boldsymbol{\nu}_k \quad (3c)$$
$$\boldsymbol{x}_k = \boldsymbol{x}(t_k) \quad (3d)$$

where $\boldsymbol{Q} \in \mathbb{R}^{n_x \times n_x}$ and $\boldsymbol{R} \in \mathbb{R}^{n_y \times n_y}$ are symmetric positive semi-definite weighting matrices and the system uncertainty is denoted by $\boldsymbol{w}_k \sim \mathcal{N}(0, \boldsymbol{Q}_k)$. The output function $\boldsymbol{h}(\cdot)$ maps the system states to the measurements $\boldsymbol{y}_k$ where $\boldsymbol{\nu}_k \sim \mathcal{N}(0, \boldsymbol{R}_k)$ is the measurement noise. The measurement noise is an independent Gaussian noise with diagonal covariance given by:

$$\boldsymbol{R}_k = \mathrm{diag}(\sigma_{p_x}^2, \sigma_{p_y}^2, \sigma_{p_z}^2, \sigma_{\omega_x}^2, \sigma_{\omega_y}^2, \sigma_{\omega_z}^2, \sigma_{a_x}^2, \sigma_{a_y}^2, \sigma_{a_z}^2) \quad (4)$$

where $\sigma_p, \sigma_\omega$ and $\sigma_a$ are the standard deviation of the noise on the position, body rate and linear acceleration measurements respectively.

To investigate the performance of the GP augmented MHE we compare the state estimation performance to MHEs with kinematic model and a dynamic model. We extend the studies to analyze the performance of these MHEs with a varying payload mass where the GP augmented MHE and the MHE with the dynamic model estimate this unknown parameter online. The formulation of these MHEs only differ in their respective models and the formulation of these are presented in the following subsections.

These models were realized in discrete time steps $\delta t$ utilizing explicit $4^{th}$ order Runge-Kutta method. The MHE was formulated using ACADOS [17], and CasADi [18] as a multiple shooting problem solving sequential quadratic program in a real-time iterating scheme.

### A. MHE with Kinematic Model

MHE with a kinematic model of the system is denoted as Kinematic-MHE (K-MHE). It is formulated such that $p_{WB}$, $\omega_B$ and $a_B$ are modeled as the measurements in the MHE. The kinematic model utilized in this MHE is given by the following equation.

$$\begin{bmatrix} \dot{p}_{WB} \\ \dot{q}_{WB} \\ \dot{v}_{WB} \\ \dot{\omega}_B \\ \dot{a}_B \end{bmatrix} = f_{kin}(x, u) = \begin{bmatrix} v_{WB} \\ q_{WB} \cdot \begin{bmatrix} 0 \\ \omega_B/2 \end{bmatrix} \\ q_{WB} \odot a_B + g_W \\ 0 \\ 0 \end{bmatrix} \quad (5)$$

$$x = \begin{bmatrix} p_{WB} & q_{WB} & v_{WB} & \omega_B & a_B \end{bmatrix}^\mathsf{T}$$
$$y = \begin{bmatrix} p_{WB} & \omega_B & a_B \end{bmatrix}^\mathsf{T}$$

### B. MHE with Dynamic Model

MHE with a dynamic model of the system is denoted as Dynamic-MHE (D-MHE). It is formulated such that $p_{WB}$ and $\omega_B$ are given as measurements and $f_{thrust}$ as an input to the system to calculate $a_B$ as formulated in (1). The model utilized in this MHE is given by the following equation.

$$\begin{bmatrix} \dot{p}_{WB} \\ \dot{q}_{WB} \\ \dot{v}_{WB} \\ \dot{\omega}_B \end{bmatrix} = f_{dyn}(x, u) = \begin{bmatrix} v_{WB} \\ q_{WB} \cdot \begin{bmatrix} 0 \\ \omega_B/2 \end{bmatrix} \\ q_{WB} \odot a_B + g_W \\ 0 \end{bmatrix} \quad (6)$$

$$x = \begin{bmatrix} p_{WB} & q_{WB} & v_{WB} & \omega_B \end{bmatrix}^\mathsf{T}$$
$$u = \begin{bmatrix} f_{thrust} \end{bmatrix} \quad y = \begin{bmatrix} p_{WB} & \omega_B & u \end{bmatrix}^\mathsf{T}$$

### C. Data-Based MHE

MHE with GPs complementing the nominal thrust model, to improve the state estimates of the agile quadrotor, is denoted as GP Augmented MHE (GP-MHE). It is formulated similar to D-MHE with an additional measurement state of $_B\hat{a}_e$ computed using the GPs to predict the acceleration error of the nominal model. The model error is assumed to be in the subspace spanned by $B_d$. The GP augmented model of the system is given by the following equation.

$$\begin{bmatrix} \dot{p}_{WB} \\ \dot{q}_{WB} \\ \dot{v}_{WB} \\ \dot{\omega}_B \\ _B\dot{\hat{a}}_e \end{bmatrix} = f_{GP}(x, u) = \begin{bmatrix} v_{WB} \\ q_{WB} \cdot \begin{bmatrix} 0 \\ \omega_B/2 \end{bmatrix} \\ q_{WB} \odot a_B + g_W \\ 0 \\ 0 \end{bmatrix} + B_{dB}\hat{a}_e$$

$$x = \begin{bmatrix} p_{WB} & q_{WB} & v_{WB} & \omega_B & _B\hat{a}_e \end{bmatrix}^\mathsf{T}$$
$$u = \begin{bmatrix} f_{thrust} \end{bmatrix} \quad y = \begin{bmatrix} p_{WB} & \omega_B & u & _B\hat{a}_e \end{bmatrix}^\mathsf{T} \quad (7)$$

As MHE keeps a record of the previous measurements, repeated computation of the model errors in the estimation horizon is unnecessary. Only a single point prediction of the current measurements is required, minimizing the computational burden of the GP.

### D. MHE with Payload Mass Estimation

MHE with parameter estimation for D-MHE and GP-MHE are formulated with their original respective models, with the addition of the payload mass term $m_p$ in the acceleration formulation:

$$a_B = \begin{bmatrix} 0 \\ 0 \\ f_{thrust}/(m + m_p) \end{bmatrix} \quad (8)$$

The unknown parameter $m_p$ is formulated as a single-degree of freedom to the optimization problem and constrained on the lower bound by 0. In this formulation the parameter uncertainty is only considered in the lower diagonal block of the new arrival cost weighting matrix $P_{k-N}^{-1}$ such that

$$P_{k-N}^{-1} = \begin{bmatrix} Q_{k-N}^{-1} & \\ & Q_p^{-1} \end{bmatrix} \quad (9)$$

where $Q_p$ is the weighting matrix of the unknown parameter. The new MHE problem with parameter estimation is formulated as follows:

$$\min_{x, w, m_p} \sum_{i=k-N}^{k} \|y_i - \hat{y}_i\|^2_{R_k^{-1}} + \|w_i\|^2_{Q_k^{-1}} \quad (10a)$$
$$+ \left\| \begin{matrix} x_{k-N} - \bar{x}_{k-N} \\ \hat{m}_p - m_p \end{matrix} \right\|^2_{P_{k-N}^{-1}}$$

subject to $\quad x_{k+1} = f_{RK4}(x_k, m_p, u_k) + w_k \quad (10b)$
$$y_k = h(x_k, u_k) + \nu_k \quad (10c)$$
$$m_{p_{min}} \leq m_p \leq m_{p_{max}} \quad (10d)$$
$$x_k = x(t_k) \quad (10e)$$

## IV. GAUSSIAN PROCESS REGRESSION

Like majority of the supervised machine learning algorithms, GP attempts to define the relationship between the inputs and the outputs of a given set of training points [19–22]. Here we utilize GP to identify the unknown dynamics of the system $d : \mathbb{R}^{n_z} \to \mathbb{R}^{n_c}$ from a set of inputs $z \in \mathbb{R}^{n_z}$ and outputs $c \in \mathbb{R}^{n_c}$:

$$c = d(z) + w_d \quad (11)$$

where the process noise $w_d \sim \mathcal{N}(0, \Sigma)$ is independent and identically distributed Gaussian noise with diagonal covariance $\Sigma = \text{diag}([\sigma_1^2, ..., \sigma_{n_d}^2])$. This allows each dimension of $c$ to be modeled independently with individual 1-dimensional GPs [23]. Given the training set $\{z, c\}$ and the test point $z_*$, the mean and the variance function of the GP is given by:

$$\mu(z_*) = k_*^\mathsf{T} K^{-1} c \quad \Sigma_{\mu k} = k_{**} - k_*^\mathsf{T} K^{-1} k_* \quad (12)$$
$$\text{with} \quad K = \kappa(z, z) + \sigma_n^2 I$$
$$k_* = \kappa(z, z_*) \qquad k_{**} = \kappa(z_*, z_*)$$

where $k_{**}$ denotes the variance of the test point, $k_*$ denotes the covariance between the training samples and the test point, and $K$ denotes the covariance matrix between the training points, also known as the Gram matrix. In this paper, we compute these (co)-variance using the Radial Basis Function (RBF) given by:

$$\kappa(z_i, z_j) = \sigma_f^2 \exp\left(-\frac{1}{2}(z_i - z_j)^\intercal L^{-2}(z_i - z_j)\right) + \sigma_n^2 \tag{13}$$

where $z_i$ and $z_j$ represent the data features, $L$ denotes the diagonal length scale matrix, and $\sigma_f$ and $\sigma_n$ represent the data and prior noise variance respectively [24]. The variables $L$, $\sigma_f$ and $\sigma_n$ shape the response of the GP regression model. During the model training process, these variables are optimized to identify the regression of the training samples. The computational complexity of the GP prediction is given by $\mathcal{O}(n^3)$ where $n$ is the number of training samples [20, 23, 25]. This motivates the use of approximation methods to handle the memory requirements and the computational demands of the traditional GP formulations [22, 23, 25].

In this paper, we implement the sparse approximation method introduced in [25], where the computational complexity is minimized by reducing the number of training points. This approach analyzes the posterior to compute the *effective prior* using another GP to encapsulate the behavior of the training samples in to a subset of $m$ number of inducing points [25]. Therefore, rather than approximating the inference, the GP is reinterpreted as an exact inference with an approximated prior. The computational complexity of the sparse GP is now given by $\mathcal{O}(m^3)$ where $m \ll n$ [25].

### A. Data Collection and Model Learning

The training data points were collected using a quadrotor with a K-MHE. At each sample time the acceleration measurement $a_B$, the estimated orientation $\hat{q}_{WB}$, and the collective thrust $f_{thrust}$ were recorded to calculate $\hat{a}_B$ using (1). The acceleration error of the dynamic model was computed by:

$$_B a_e = a - \hat{a} \tag{14a}$$
$$\text{where} \quad a = a_B + \hat{q}_{WB}^{-1} \odot g_W \tag{14b}$$
$$\hat{a} = \hat{q}_{WB}^{-1} \odot (\hat{q}_{WB} \odot \hat{a}_B + g_W) \tag{14c}$$

The GP models were trained such that the acceleration measurements $a_B$ were mapped to the body frame acceleration disturbances $_B a_e$. The unmodelled dynamics in each axis were modeled independently to minimize the required inducing points of the GP models. Therefore, three individual GP models $\mu_{ax}, \mu_{ay}, \mu_{az}$ were developed. The GP predictions are denoted as follows

$$_B\hat{a}_e = \boldsymbol{\mu}_a(_B a_m) = \begin{bmatrix} \mu_{ax}(_B a_x) \\ \mu_{ay}(_B a_y) \\ \mu_{az}(_B a_z) \end{bmatrix}$$
$$\boldsymbol{\Sigma}_\mu(a_B) = \text{diag}\left(\begin{bmatrix} \Sigma_{ax}(_B a_x) \\ \Sigma_{ay}(_B a_y) \\ \Sigma_{az}(_B a_z) \end{bmatrix}\right) \tag{15}$$

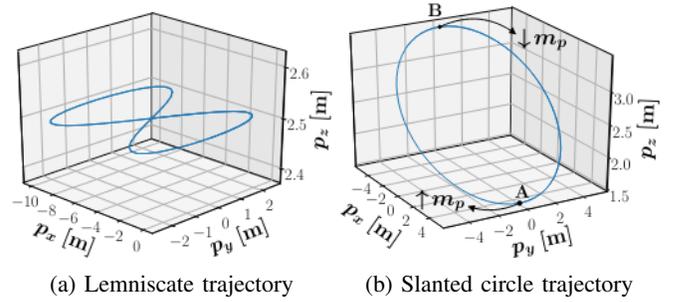

(a) Lemniscate trajectory  (b) Slanted circle trajectory

Fig. 3: Quadrotor trajectories utilized in Simulation studies.

## V. SIMULATION STUDIES

### A. Simulation Setup

We first evaluate the the performance of the state estimators in a Robot Operating System (ROS) Gazebo [26] environment utilizing RotorS [27] simulator package to simulate the AscTec Hummingbird quadrotor. The sensors simulated were GPS and IMU and the measurements were received at 100Hz with cascaded zero-mean Gaussian distributed noise at varying levels as indicated in Table I. The simulation studies were conducted on laptop with 16GB of RAM, 10th Generation Intel Core i7-10750H and NVIDIA GeForce RTX 2070 8GB GDDR6.

In the simulations studies, we compare the state estimation performance of the MHEs with varying models as described in Section III: K-MHE, D-MHE and GP-MHE. To investigate the performance of the proposed estimation method, we analyze the quadrotor executing two different trajectories illustrated in Fig. 3. The lemniscate trajectory is defined by $[x(t) = 5\cos(\sqrt{2}t) - 5,\, y(t) = 5\sin(\sqrt{2}t)\cos(\sqrt{2}t),\, z(t) = 2.5]$. The slanted circle trajectory is defined by $[x(t) = 5\cos(t),\, y(t) = 5\sin(t),\, z(t) = -\cos(t) + 2.5]$. Both trajectories accelerates from a hovering state, reaching a peak velocity of $11.3 ms^{-1}$ and $8.7 ms^{-1}$ respectively and decelerates back down to $0 ms^{-1}$. The slanted circle trajectory was utilized to imitate a quadrotor, weighing $m = 1$kg, transporting an object weighing 300g, from point A to point B to investigate the state estimation performance with a varying parameter.

### B. Trained GP Regression Models

Each GP models were trained for every corresponding noise levels and trajectory. The GP models mapping the accelerometer measurements to the acceleration error at measurement noise level III are presented in Fig. 4. These models are sparse GPs trained with 50 inducing points derived from dense GPs. The GPs on the left column are trained from data points collected by a quadrotor executing the lemniscate trajectory. The GPs on the right column are trained from data

TABLE I: Standard deviation of the measurement noise at varying levels

|  | $\sigma_p$ | $\sigma_\omega$ | $\sigma_a$ |
|---|---|---|---|
| Noise Level I | 0.007 | 0.40 | 0.007 |
| Noise Level II | 0.5 | 0.86 | 0.01 |
| Noise Level III | 1 | 1.72 | 0.1 |

points collected by a quadrotor executing the slanted circle trajectory.

It was found that within the range of 20 to 60 inducing points, the sample size of the GPs had little to no effect in the performance and the required computation time of the GP-MHE. This can be explained by the near linear relationship of the GPs in the $x$ and $y$ axis and the use of sparse GPs. As the general trend of the dataset is learned by a larger GP and the sparse GP is trained using effective priors. A GP trained using 20 effective priors can produce similar regression as a GP trained from 60. Furthermore, as only a single point prediction is required at every time step, the differences in the additional computational time of these GPs are negligible.

*C. Simulation Results of the GP augmented MHE*

We first investigate the trade-off between the optimization time and the estimation performance with respect to the number of estimation nodes in the MHE. The Root Mean Squared Error (RMSE) of $q$ and $v$ state estimates and the optimization time executing the lemniscate trajectory at measurement noise level III are visualized in Fig. 5. The computational complexity follows a near linear relationship with respect to the estimation horizon length, while the estimation performance plateaus around 45 nodes. From this we chose the MHEs to be formulated with estimation horizon of 0.5 seconds with 50 nodes, corresponding to $3-4.5$ms of optimization time. It can also be noted that the GP only adds approximately 2ms of additional computational time to the D-MHE regardless of the number of estimation nodes.

The stability of the MHEs to the measurement noise were investigated on the lemniscate trajectory at three levels of measurement noise. Table II summarizes the state estimation performance of the three estimators.

Due to the limited knowledge of the system, D-MHE performed poorly at all noise levels. It is to be noted that the external disturbances in real-time application will exceed the aerodynamic effects simulated in these studies, resulting in further inaccurate estimates by the D-MHE. With the GP model corrections the performance was significantly improved. The orientation states were improved by 73%, 41%; and 30%, and the velocity state estimates were improved by 64%, 23%, and 15%, at noise levels I, II and III respectively.

At noise level I and II, K-MHE produced the most accurate state estimates, closely followed by the GP-MHE. It was expected that the K-MHE outperform GP-MHE with low measurement noise where the sensor measurements can accurately describe the dynamics of the system. At higher noise levels, the quality of the measurements are deterred, hence the slight improvement of GP-MHE over K-MHE. At noise level III the RMSE of the orientation state estimates of K-MHE and GP-MHE are $7.049°$ and $6.791°$ respectively. The RMSE of the velocity state estimates of K-MHE and GP-MHE are of similar level: $1.000 ms^{-1}$ and $0.987 ms^{-1}$ respectively. This can be explained by the robustness of the GPs to Gaussian distributed noise, thus it can provide more accurate dynamics to the MHE over direct measurement of the accelerometer.

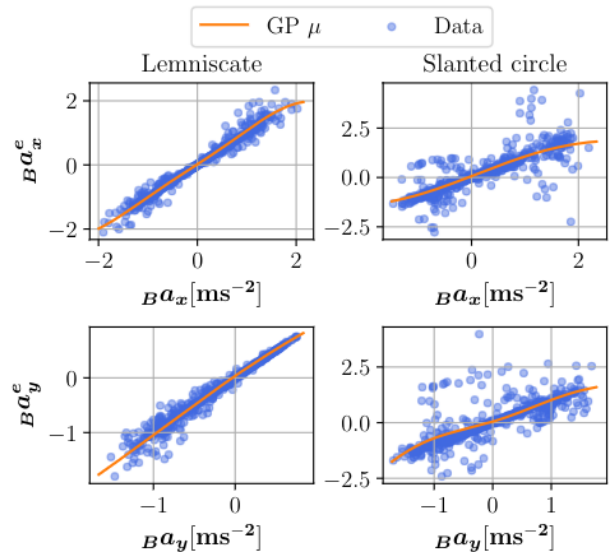

Fig. 4: Sparse GP models with 50 inducing points mapping the body frame acceleration error observed on the quadrotor in Noise Level III simulation.

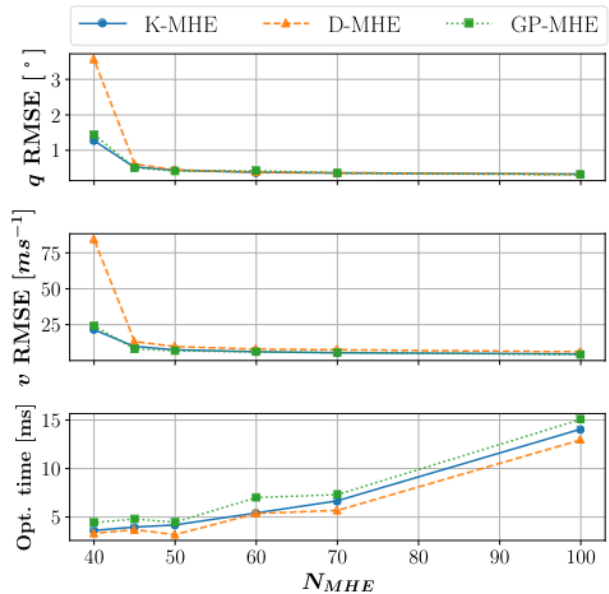

Fig. 5: Trade-off between the number of estimation nodes, estimation performance in $q$ and $v$ states and optimization time at Noise level III simulation.

TABLE II: Comparison of state estimation results of MHEs at varying noise levels in simulation.

|  |  | Model RMSE | | |
| --- | --- | --- | --- | --- |
|  |  | K-MHE | D-MHE | GP-MHE |
| Noise Level I | p [m] | 0.059 | 0.198 | 0.056 |
|  | q[°] | 1.540 | 5.907 | 1.596 |
|  | v[m/s] | 0.168 | 0.625 | 0.223 |
| Noise Level II | p [m] | 0.200 | 0.276 | 0.240 |
|  | q[°] | 3.157 | 7.163 | 4.232 |
|  | v[m/s] | 0.473 | 0.781 | 0.604 |
| Noise Level III | p [m] | 0.420 | 0.447 | 0.413 |
|  | q[°] | 7.049 | 9.708 | 6.791 |
|  | v[m/s] | 1.000 | 1.159 | 0.987 |

## D. Simulation Results of the GP augmented MHE with online parameter estimation

We investigate the state estimation performance with a varying parameter $m_p$ while executing the slanted circle trajectory. The mass of the quadrotor was increased by 300g at point **A** and reduced back down to its original mass at point **B**, where the original mass of the quadrotor is 1kg. This was repeated for the entire duration of the trajectory as the quadrotor accelerated to its maximum velocity and decelerated back down to its hovering state. The MHE with the payload mass estimation were formulated as discussed in Section III-D to include the additional unknown parameter.

The performance of the MHEs with and without the additional payload are presented in Table III. The estimation of the payload mass by the MHEs are presented in Fig. 6. It can be noted that due to the extra estimation parameter and its constraints, the required computation time has increased by $2.1 - 2.7 ms$; however, it is still less than its sampling rate of $10 ms$.

It is evident that the D-MHE and GP-MHE can successfully estimate the varying mass of the quadrotor. This is further supported by the insignificant differences between the state estimation performance of 0g and 300g payload mass. Despite accurate parameter estimation, D-MHE performed poorly due to the weak dynamic model. By introducing the GP models to the MHE, the $q$ and $v$ state estimates improved by 40% and 19% respectively, performing at similar level to K-MHE. This shows that the GP-MHE can successfully handle varying parameters while still providing accurate state estimations of the system. This can be advantageous over K-MHE as knowledge of an unknown system parameter can be improve the trajectory tracking performance of a controller.

## VI. EXPERIMENTAL STUDIES

### A. Experimental Setup

We finally conduct comparative analysis of the MHEs on a real-time NeuroBEM dataset from the University of Zurich to further validate the performance of the GP-MHE [28]. The dataset contains Vicon and onboard measurements of the agile quadrotor flights. For further details on data collection please refer to [28]. The trajectory of the flight data that was utilized is presented in Fig. 7. The recorded flight reached a maximum velocity of $15 m/s$ with a maximum acceleration of $42.5 m/s^2$. Firstly, we tested the performance of the MHEs with the given dataset to simulate the state estimation performance in an indoor setting with accurate position measurements. The MHE of the *indoor* experiment was formulated with the position measurements from Vicon and inertial measurements from the onboard IMU. Secondly, to imitate the inaccurate GPS measurements in an outdoor environment we cascade a Gaussian distributed noise of $\sigma_p = 1m$ to the Vicon position measurements. The MHE of the *outdoor* experiment was formulated with the position measurements from Vicon with the added noise and the inertial measurements from the onboard IMU.

TABLE III: State estimation performance of the quadrotor with changing mass in simulation with Noise Level III and the optimization time of each algorithm.

| Payload Mass [kg] | | Model RMSE | | |
|---|---|---|---|---|
| | | K-MHE | D-MHE | GP-MHE |
| 0 | p [m] | 0.393 | 0.431 | 0.389 |
| | q[°] | 6.819 | 11.565 | 6.683 |
| | v[m/s] | 0.903 | 1.185 | 0.915 |
| | Opt. dt [ms] | 4.309 | 3.164 | 4.464 |
| 0.3 | p [m] | 0.401 | 0.473 | 0.437 |
| | q[°] | 6.622 | 10.593 | 6.336 |
| | v[m/s] | 0.951 | 1.262 | 1.019 |
| | Opt. dt [ms] | 4.264 | 5.274 | 7.116 |

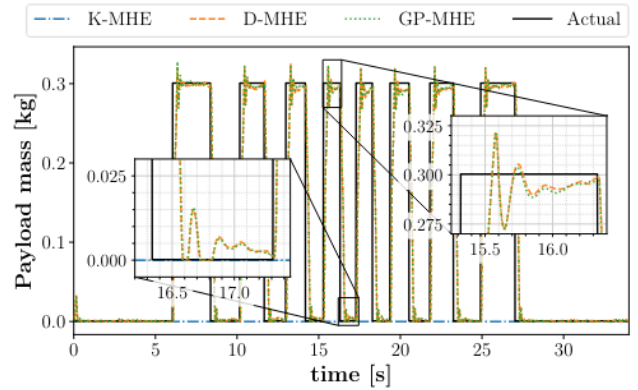

Fig. 6: Comparison of payload mass estimation result of the MHEs in simulation of Noise Level III.

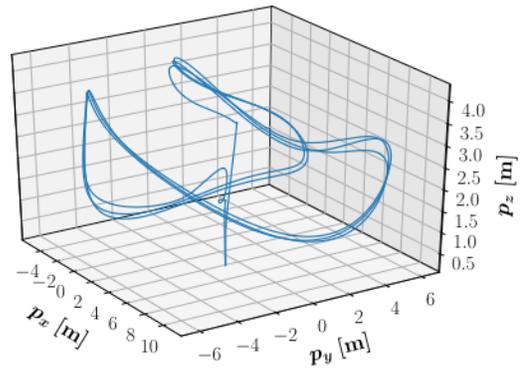

Fig. 7: Trajectory executed by an agile quadrotor in the Experimental Studies.

### B. Experimental Results

The GP models mapping the accelerometer measurements to the model errors recorded from a real-time quadrotor platform is presented in Fig. 8. These data points are calculated from the quadrotor executing the drone racing trajectory shown in Fig. 7. The visualized models are sparse GPs trained with 50 inducing points. The left is the model trained with the data collected in an indoor setting and the right is the model trained with the data collected in an outdoor setting.

The RMSE of the state estimation performance of the three MHEs are tabulated in Table IV. Due to the increased agility and aggressiveness of the trajectory the errors seen on the experimental results are higher than those from the simulation results. The RMSE results are an average of three experiments as the results may vary due to the stochastic behavior from the cascaded noise for the outdoor setting. In an indoor setting, the $q$ state estimates of GP-MHE improved by 26% and 5.6% compared to D-MHE and K-MHE respectively. While there was a minute decrease in the accuracy of the $v$ state estimates. In an outdoor setting, the $q$ state estimates improved by 14.5% and 4% compared to D-MHE and K-MHE respectively. The $v$ state estimates increased by 4.2% compare to D-MHE however the accuracy decreased by 1.8% compared to K-MHE.

The enhancements in the state estimation performance of the GP-MHE is evident in the results with significant improvements in the estimation of the angular position of the quadrotor. The particular improvement in the $q$ state estimates can be explained by the fact that the dynamic model assumes there are no accelerations in the $x$ and $y$ directions in the body frame from the rotor thrusts. In the D-MHE, these are compensated with the $q$ states to capture those movements, with the consequences of increased error in the angular positions. Furthermore, the authors of [7, 28] explains that the main source of disturbance experienced by the given quadrotor is the rotor drag due to its power and the compactness of its design. Where in GP-MHE, the GP models are trained with acceleration errors that are coupled with these angular position estimates, it is able to capture these disturbances and correct for these model errors.

## VII. CONCLUSION

In this work, we have presented a data-based MHE method for agile quadrotors. The dynamic model of the system is augmented with GPs to provide model corrections for unmodelled aerodynamic effects. The training points were collected from on-board sensors rather than ground truth data simplifying the data collection process. The GP models were then trained to predict the acceleration error of the nominal model given its accelerometer measurements. As MHE keeps a record of its past measurements, the computational burden of the GP could be minimized by only requiring single point prediction of the current measurements. The simulation and experimental studies have demonstrated a significant improvement in the state estimation performance compared to traditional dynamic model-based estimation methods. Furthermore, GP-MHE achieved a similar level of performance to K-MHE while also offering the added capability of estimating unknown model parameters online. This valuable feature can complement the controller and enhance the closed-loop tracking performance of the quadrotor.

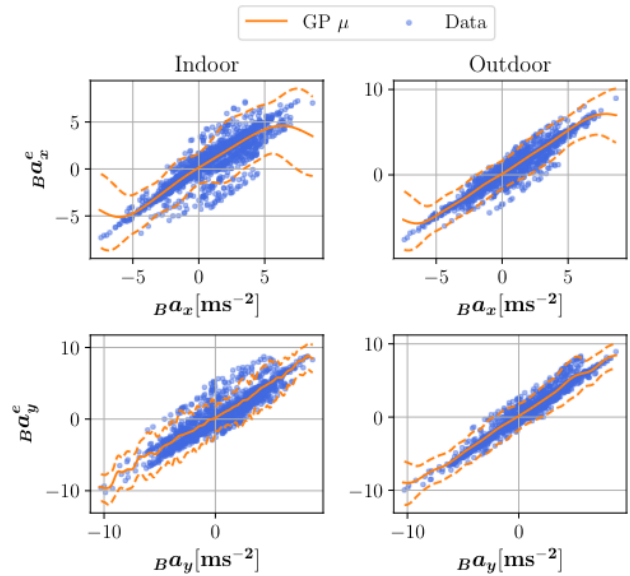

Fig. 8: The GP models mapping the relationship between the acceleration measurements to model errors observed on a real quadrotor platform executing a drone racing trajectory.

TABLE IV: Experimental results of the MHEs estimating the states of a real-world quadrotor executing drone racing trajectory.

|  |  | Model RMSE | | |
|---|---|---|---|---|
|  |  | K-MHE | D-MHE | GP-MHE |
| INDOOR | p [m] | 0.072 | 0.080 | 0.076 |
|  | q[°] | 16.348 | 20.927 | 15.436 |
|  | v[m/s] | 5.828 | 5.810 | 5.940 |
| OUTDOOR | p [m] | 0.395 | 0.415 | 0.413 |
|  | q[°] | 18.665 | 20.972 | 17.921 |
|  | v[m/s] | 5.869 | 6.241 | 5.981 |